\documentclass{article}
\usepackage{ijcai24}

\usepackage{times}
\usepackage{soul}
\usepackage{url}
\usepackage[hidelinks]{hyperref}
\usepackage[utf8,cp1250]{inputenc}
\usepackage[small]{caption}
\usepackage{graphicx}
\usepackage{amsmath}
\usepackage{amsthm}
\usepackage{booktabs}
\usepackage{algorithm}
\usepackage{algorithmic}
\usepackage[switch]{lineno}
\usepackage{tikz}

\usepackage{hyperref}
\usepackage{color}

\def\our{ImplicitDeepfake}

\usepackage{natbib}

\def\F{\mathcal{F}}

\def\G{\mathcal{G}}
\def\N{\mathcal{N}}
\def\m{\mathrm{m}}
\def\x{ {\bf x} }
\def\F{\mathcal{F}}
\def\I{\mathcal{I}}

\title{Deepfake for the Good: Generating Avatars through Face-Swapping with \\ Implicit Deepfake Generation}

\author{
Georgii Stanishevskii$^1$$^*$
Jakub Steczkiewicz$^1$$^*$\and
Tomasz Szczepanik$^1$$^*$\and\\
S\l{}awomir Tadeja$^2$\and
Jacek Tabor$^1$ \and
Przemys\l{}aw Spurek$^1$\\
\affiliations
$^*$Equal contribution\\
$^1$Jagiellonian University, Faculty
of Mathematics and Computer Science, Cracow, Poland\\
$^2$Department of Engineering, University of Cambridge,
Cambridge, UK\\
\emails
przemyslaw.spurek@uj.edu.pl
}

\begin{document}

\maketitle

\begin{abstract}   
Numerous emerging deep-learning techniques have had a substantial impact on computer graphics. Among the most promising breakthroughs are the rise of Neural Radiance Fields (NeRFs) and Gaussian Splatting (GS). NeRFs encode the object's shape and color in neural network weights using a handful of images with known camera positions to generate novel views. In contrast, GS provides accelerated training and inference without a decrease in rendering quality by encoding the object's characteristics in a collection of Gaussian distributions. These two techniques have found many use cases in spatial computing and other domains. On the other hand, the emergence of deepfake methods has sparked considerable controversy. Deepfakes refers to artificial intelligence-generated videos that closely mimic authentic footage. Using generative models, they can modify facial features, enabling the creation of altered identities or expressions that exhibit a remarkably realistic appearance to a real person. Despite these controversies, deepfake can offer a next-generation solution for avatar creation and gaming when of desirable quality. To that end, we show how to combine all these emerging technologies to obtain a more plausible outcome. Our \emph{\our{}} uses the classical deepfake algorithm to modify all training images separately and then train NeRF and GS on modified faces. Such simple strategies can produce plausible 3D deepfake-based avatars.
\end{abstract}

\section{Introduction}

\begin{figure}[!t]
    \begin{center}
    {\includegraphics[width=0.5\textwidth]{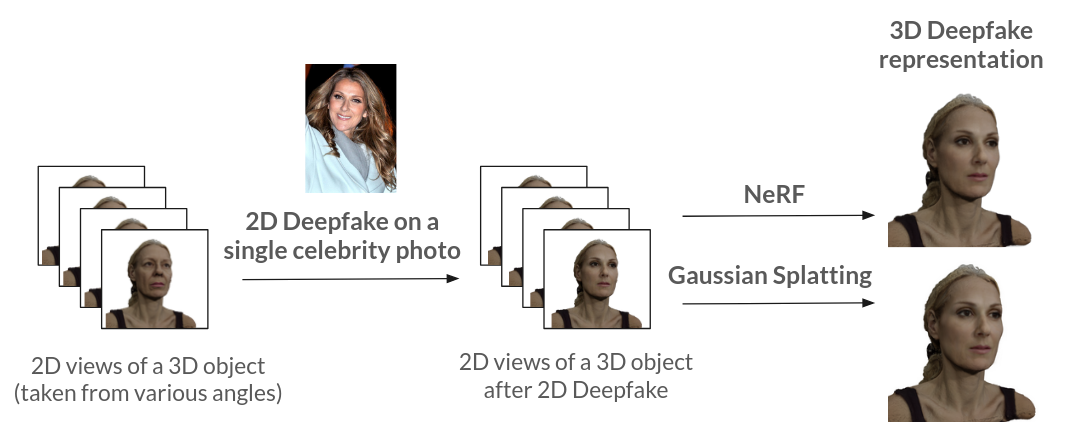} }


    \end{center}
    \caption{ Thanks to neural rendering, we can aggregate information from 2D images to produce novel views of 3D objects (see left-hand column {\em Neural rendering on original images}). In \our{}, we utilize 2D DeepFake with the help of a single image and then neural rendering to effectively obtain 3D DeepFake (see the right column {\em Neural rendering on modified images}).}
    \label{fig:merf_appro}
\end{figure}

Manipulating images and videos have a long history and spikes many controversies when misused \cite{zhang2022deepfake}. Moreover, a number of dedicated to this task software tools and packages like Adobe
Photoshop\footnote{\url{https://www.adobe.com/products/photoshop.html}} or Adobe Lightroom\footnote{\url{https://www.adobe.com/products/photoshop-lightroom.html}} have been developed and commercially available, further contributing towards their prevalence \cite{maares2021labour}. At the same time, while controversial, these technologies also hold potential for their positive use with emerging applications in healthcare \cite{DeepfakeHealthcare2024} (e.g. in plastic surgery practice \cite{crystal_photographic_2020}), scientific research (e.g. personas generation for human-computer interaction (HCI) research \cite{KAATE2023103096, KAATE2024100031, DeepFakePersonasinHCI2023}) or entertainment (e.g. generation of video game content such as animations or graphics \cite{lundberg_potential_2024}). One such domain is the potential to generate plausible digital avatars \cite{DeepFakeAvatars2024}.
The latter is becoming increasingly more important due to the rapid growth of various virtual environments and online platforms.
This necessitates a comprehensive understanding of human behavior when immersed within these digital spaces to inform the design of ethical and effective virtual environments \cite{SocialDiversityAvatars2023, DeepFakePersonasinHCI2023} in which the role of avatars is considered to grow in the future \cite{SocialDiversityAvatars2023}. Here, the realism of generated avatars can have a profound impact on user acceptance and perception \cite{KAATE2023103096, KAATE2024100031, DeepFakePersonasinHCI2023}.

In this context, realistic facial feature modification that allows for a convincing avatar creation, i.e. resembling someone else appearance to a great extent, remains a non-trivial task that can be supported with emerging deepfake technology~\cite{waseem2023deepfake}. The deepfake term is a combination of {\em deep learning } and {\em fake} referring to manipulated media content created using machine learning and artificial neural networks~\cite{waseem2023deepfake}. 

Various deepfake methods employ deep generative models like autoencoders \cite{tewari2018high} or generative adversarial networks (GANs) \cite{kumar2020detecting} for examining the facial characteristics and mimics of a given individual. Such analyses enable the creation of manipulated facial pictures that imitate comparable expressions and motions \cite{pan2020deepfake}.

The availability of user-friendly tools, including DeepFaceLab \cite{liu2023deepfacelab}, smartphone applications such as Zao\footnote{\url{https://zaodownload.com}} and FakeApp\footnote{\url{https://www.malavida.com/en/soft/fakeapp/}} has simplified the use and fostered the adoption of deepfake by nonprofessionals helping them to swap faces with any target person seamlessly for any desired purpose. Deepfake generation and detection are characterized by intense competition, with defenders (i.e., detectors) and adversaries (i.e. generators) continuously trying to outdo each other. A range of notable advances have been made in both areas in recent years \cite{turek2019media}.

To extend this growing body of research, we present in this paper \our{}\footnote{\url{https://quereste.github.io/implicit-deepfake/} \url{https://github.com/quereste/implicit-deepfake}}--a first model that produces a 3D deepfake. To obtain real word 3D objects, we use novel, machine learning-based computer graphics techniques from the field of neural rendering such as \textit{Neural Radiance Fields} (NeRFs)~\cite{mildenhall2020nerf} and \textit{Gaussian Splatting} (GS) \cite{kerbl20233d}. 
These two rendering methods, i.e. NeRFs and GS, can be considered converters from 2D to 3D objects, where as the input, we provide 2D images with viewing positions, and as output, we can obtain 3D objects. In the context of 3D deepfake, we can use a classical model dedicated to 2D images to produce input to the NeRF-based model. \our{} applies 2D deepfake for all 2D images separately and then adds to NeRF architecture as shown in Fig.~\ref{fig:merf_appro}. 
In practice, deepfake technology produces consistent image swapping, and we can directly use neural rendering to produce 3D faces. \our{} can generate a 3D avatar from just a single image of the person used for the deepfake.

\begin{figure}[th]
	\centering

    \includegraphics[width=0.5\textwidth, trim=0 0 0 0, clip]{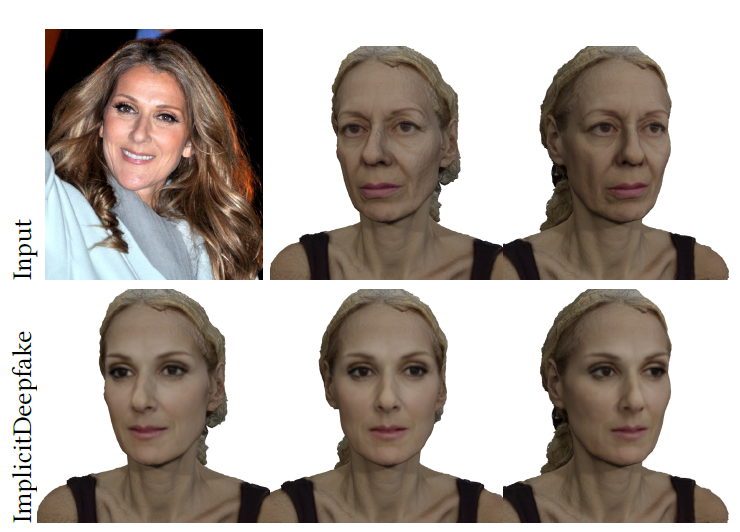}  

\caption{In our paper, we present \our{} which uses single image and universal 3D model on input and produces 3D avatar.}
\label{fig:compape_1}
\end{figure}

Our avatars can also be edited with straightforward text prompts by applying diffusion models. Diffusion models \cite{ho2020denoising} are types of generative models that can generate images from text prompts. These models incrementally corrupt the training data with Gaussian noise and subsequently learn to reconstruct the data by reversing this noise addition. The Diffusion Model can generate data post-training by feeding randomly sampled noise through the trained denoising mechanism. One of the most powerful features of the model is that we can condition the results with text and images \cite{zhang2023addingconditionalcontroltexttoimage}. In our \our{} framework, we apply diffusion modification to the input images and use the NeRF or GS algorithm to reconstruct the 3D avatar.
Thus, \our{} can offer the next-generation, high-quality approach for plausible avatar creation using a single image and text prompt.  

In summary, our contributions are as follows:
\begin{itemize}
    \item We propose a new method dubbed~\our{}, which produces a 3D avatar from a single image and a text prompt.
    \item \our{} combines neural rendering procedure with 2D deepfake and diffusion models.
    \item 
    We show that the 2D deepfake and 2D diffusion models can produce consistent images for NeRF and GS.
\end{itemize}

\section{Related Works}

This section discusses various proposed methods for face swapping in images and videos. In most cases, authors use large generative models like autoencoders \cite{tewari2018high} and generative adversarial networks (GAN) \cite{kumar2020detecting} or diffusion models \cite{ho2020denoising}.

Performing a face swap classically involves three key stages \cite{waseem2023deepfake}. First, the algorithm identifies faces in both the source and target videos. Subsequently, the method substitutes the target face's nose, mouth, and eyes with the corresponding features from the source face. The color and lighting of the candidate's facial image are modified to ensure a smooth integration of the two faces. Finally, the overlapping region undergoes match distance computation to assess and rank the quality of the merged candidate replacement.

Classical convolutional neural networks (CNNs) were also employed for generating deepfakes \cite{korshunova2017fast}. However, these methods are limited, as they can only transform individual images. Therefore, they are unsuitable for creating high-quality videos.

Moreover, Reddit\footnote{\url{https://www.reddit.com/r/deepfakes/}} platform introduced a deepfake video creation method that makes use of an autoencoder architecture. This approach involves a deepfake face-swapping autoencoder network consisting of one encoder and two decoders. Throughout the training phase, the encoder and the two decoders share parameters. The autoencoder technique is utilized by different face-swapping applications, including DeepFaceLab \cite{liu2023deepfacelab} and DFaker\footnote{\url{https://github.com/dfaker/df}}.

The next group of generative models used in the deepfake area were GANs (Generative Adversarial Networks). For example, the Face-swap GAN (FS-GAN) \cite{nirkin2019fsgan} opts for an encoder-decoder architecture as its generator, along with antagonistic and perceptive losses, to enhance the automated coding system. Including counter losses has resulted in improved image reconstruction efficiency while using perceptual loss, which has helped align the generated face with the input image. On the other hand, the RSGAN \cite{natsume2018rsgan} utilizes two Variational Autoencoders (VAE) to produce distinct latent vector embeddings for hair and face areas. These embeddings are subsequently merged to generate a face that has been swapped. 

FSNet \cite{natsume2019fsnet} simplifies the pre- and postprocessing phases by removing their complexity. FSNet consists of two sub-networks: a VAE, which produces the latent vector for the face region in the source image, and a generator network, which combines the latent vector of the source face with the non-face components, such as hairstyles and backgrounds of the target image, thereby achieving face swapping.

The FaceShifter approach \cite{li2019faceshifter} employs a two-step solution. Initially, a GAN network is utilized to extract and flexibly merge the identities of the source and target images. Subsequently, the occluded regions are refined using the Heuristic Error Acknowledging Refinement Network (HEAR). In the study conducted by \cite{xu2022high}, StyleGAN is employed to separate the texture and appearance characteristics of facial images. This technique allows for preserving the desired target look and texture while maintaining the source identity. 

Many different approaches exist in the literature, as well as multiple GitHub repositories. Thus, it is not practical to verify all existing methods. Therefore, during experiments, we chose to use GHOST \cite{9851423}. To the best of our knowledge, \our{} is the first model that utilizes the NeRF and GS approach for generating 3D deepfakes.

\begin{figure*}
 	\centering

     \includegraphics[width=0.9\textwidth]{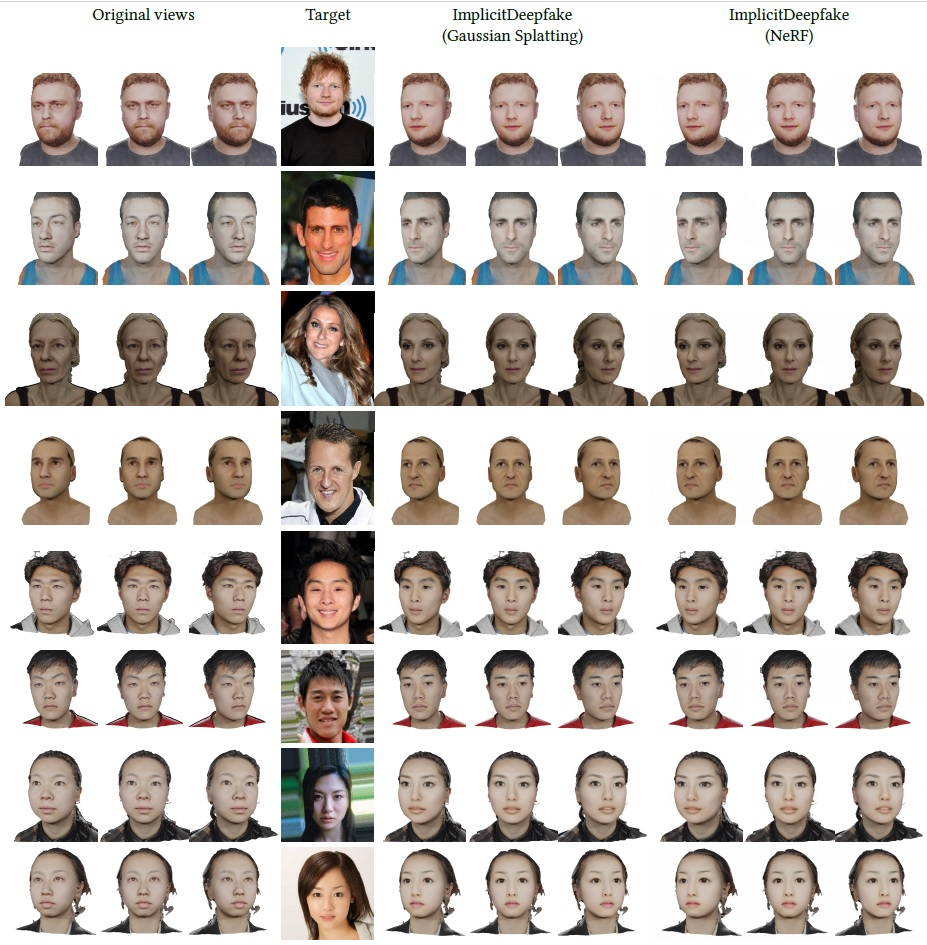} 

\caption{Comparison between \our{} trained on NeRF and GS. In the first column, we see the original input 3D avatars. Then, we present an image of the celebrity who is the target of deepfake. In the last two columns, we have the results obtained with the help of NeRF and GS. In general, GS provides more visually plausible renders.}
\label{fig:deep_fake_GS} 
\end{figure*}

\section{Method}

This section introduces our model \our{}. First, we describe NeRF and GS technology and establish a notation. Then, we describe a deepfake and diffusion model used in our approach. Finally, we present our model. 

\paragraph{Neural Radiance Field (NeRF)}

The vanilla NeRF model proposed by \cite{mildenhall2020nerf} is a neural architecture that represents 3D scenes. In NeRF, a 5D coordinate is taken as input, consisting of the spatial location $\x = (x, y, z)$ and the viewing direction $\mathbf{d} = (\theta, \psi)$. The model then outputs the emitted color $\mathbf{c} = (r, g, b)$ and the volume density $\sigma$. 

A classical NeRF model utilizes a set of images during the training process. This approach generates a group of rays, which are passed through the image. Neural networks predict color and depth information on such rays to represent a 3D object. NeRF encodes such 3D objects using a Multilayer Perceptron (MLP) network: 
$$ 
\F_{NeRF} (\x , {\bf d}; \Theta ) = ( {\bf c} , \sigma). 
$$ 


The training process involves optimizing the parameters ($\bf \Theta$) of the MLP to minimize the difference between the rendered and reference images obtained from a specific dataset. This calibration allows the MLP network to take in a 3D coordinate and its corresponding viewing direction as input and then produce a density value and color (radiance) along that direction.







The NeRF's performance limitations can be attributed to two primary factors, i.e., the inherent capacity constraints of the underlying neural network and the computational difficulty of accurately computing the intersection points between camera rays and the scene geometry. These limitations can lead to prolonged rendering times for high-resolution images, particularly for complex scenes, impeding the potential of real-time applications.

\paragraph{Gaussian Splatting (GS)}



GS model employs a collection of 3D Gaussian functions to represent a 3D scene. Each Gaussian is characterized by a set of parameters, i.e., its position (mean) specifying its center, its covariance matrix defining the shape and orientation of the Gaussian distribution, its opacity controlling the level of transparency, and its color represented by spherical harmonics (SH) \cite{fridovich2022plenoxels,muller2022instant}.


The GS represents a radiance field by optimizing all the 3D Gaussian parameters. Moreover, the computational efficiency of the GS algorithm stems from its rendering process, which leverages the projection properties of Gaussian components. This approach relies on representing the scene with a dense set of Gaussians, mathematically denoted as:

$$
\G = \{ (\N(\m_i,\Sigma_i), \sigma_i, c_i) \}_{i=1}^{n},
$$ 
where $\m_i$ is the position, $\Sigma_i$ stands for the covariance, $\sigma_i$ is the opacity, and $c_i$ are the $i$-th component's SH for color representation.


The GS optimization algorithm operates through an iterative image synthesis process and comparison with training views. The challenges of this procedure arise due to potential inaccuracies in Gaussian component placement stemming from the inherent dimensionality reduction of the 3D to 2D projection. To mitigate these issues, the algorithm incorporates mechanisms for creating, removing, and relocating Gaussian components. This enables GS to achieve visual quality comparable to NeRF's while potentially exhibiting faster training and inference times.


\begin{table*}[]
{\small
    \caption{Comparison between deepfake on test positions (camera position that is not used in training of \our{}) obtained by \our{} and direct application of deepfake. Results are calculated using faces from Fig.~\ref{fig:deep_fake_GS}.  }
\begin{center}
    \begin{tabular}{l|lllllllll}
    \multicolumn{9}{c}{PSNR $\uparrow$} \\
         & FACE 1 & FACE 2 & FACE 3 & FACE 4 & FACE 5 & FACE 6 & FACE 7 & FACE 8 & Avg. \\ \hline

NeRF & 37.89 & 38.58 & 37.95 & 41.01 & 36.86 & 37.77 & 35.91 & 35.14 & 37.64 \\ 
GS & \bf 41.58 & \bf 41.71 & \bf 42.45 & \bf 44.00 & \bf 41.25 & \bf 41.73 & \bf 39.65 & \bf 41.43 & \bf 41.73 \\  

 \hline

    \multicolumn{9}{c}{SSIM $\uparrow$} \\
\hline
NeRF & 0.97 & 0.97 & 0.98 & 0.99 & 0.98 & 0.98 & 0.98 & 0.97 & 0.98 \\ 
GS & \bf 0.99 &  \bf  0.99 &  \bf  0.99 &  \bf  0.99 &  \bf 0.99 &  \bf 1.00 &  \bf 0.99 &  \bf 0.99 &  \bf 0.99  
\\ \hline

    \multicolumn{9}{c}{LPIPS $\downarrow$ }\\ \hline
NeRF & 0.08 & 0.08 & 0.07 & 0.06 & 0.04 & 0.04 & 0.05 & 0.06 & 0.06 \\ 
GS &  \bf  0.04 &  \bf  0.04 &  \bf  0.03 &  \bf  0.03 &  \bf  0.02 &  \bf  0.01 &  \bf  0.02 &  \bf  0.02 & \bf  0.03 \\  
\hline
    \end{tabular}
    \end{center}
    \label{tab:celeb}
    }

\end{table*} 

\paragraph{Deepfake with Generative High-fidelity One Shot Transfer (GHOST)}

Numerous deepfake technologies have been developed in the literature, each with its own set of limitations. These may include errors in face edge detection, inconsistencies in eye gaze, and overall low quality, particularly when replacing a face from a single image with a video \cite{waseem2023deepfake}. An example approach to deepfake is represented by GHOST \cite{9851423}, in which the authors build on the FaceShifter \cite{li2019faceshifter} model as a starting point and introduce several enhancements in both the quality and stability of deepfake. Consequently, we decided to utilize GHOST as a deepfake component of our model. 

Let $I_s$ and $I_t$ be the faces cropped from the source and target images, respectively, and let $\hat I_{s,t}$ be the newly generated face. The architecture of GHOST consists of a few main parts. The identity encoder (U-Net \cite{ronneberger2015u}) extracts information from the source image $I_s$ and keeps the information about the identity of the source person.

\begin{figure*}[th]
	\centering

    \includegraphics[width=0.9\textwidth]{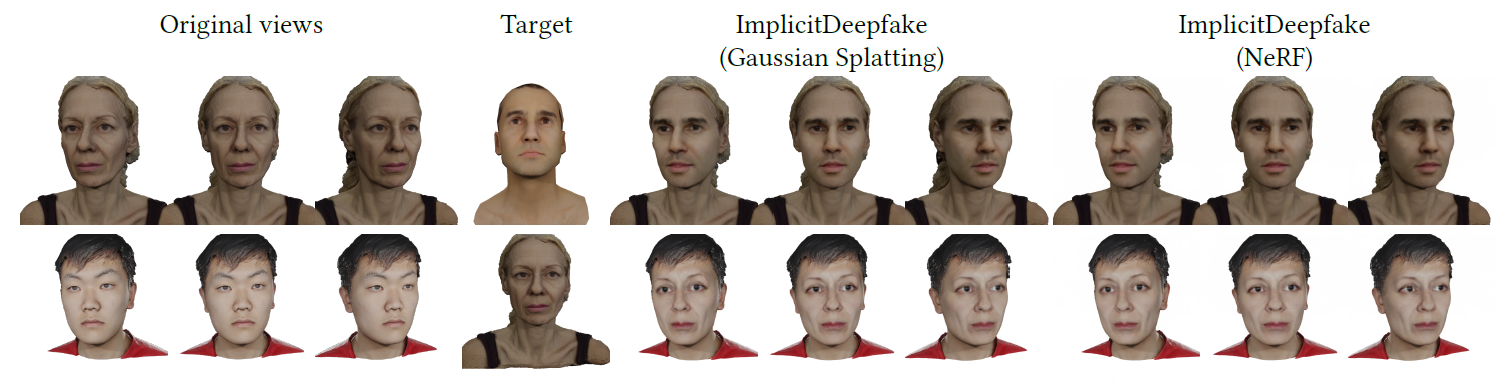} 

     \caption{\our{} returns satisfying results even when the source and target faces bear little resemblance to each other. In the following comparison, we present two versions of \our{} (NeRF- and GS-based) in a setup where the source and target are of different sexes.}
\label{fig:deep_fake_GS_1} 
\end{figure*}

The AAD generator is a model that combines the attribute vector obtained from $I_t$ and the identity vector obtained from $I_s$ successively using AAD ResBlocks. This process generates a new face $\hat I_{s,t}$ that possesses the attributes of both the source identity and the target. Moreover, using a multi-scale discriminator \cite{park2019semantic} improves the quality of output synthesis by comparing genuine and fake images.
The GHOST model uses the cost function, which consists of a few elements representing reconstruction, attributes, identity, and the GAN loss based on a multi-scale discriminator.




Our approach takes the main face shape from the source image and changes only the cropped part. Therefore, we can use deepfake and neural rendering components separately.

\paragraph{Diffusion models }

In diffusion modules, we consider the forward and backward diffusion processes. We assume that we are given a controlling sequence $\beta_t \in (0,1)$. In the forward process, we start from the point $x_0$ in the data manifold $M$, and put: 
$$
x_t=\sqrt{1-\beta_t}x_{t-1}+\sqrt{\beta_t} \varepsilon_t, \text{ where }\varepsilon_t \sim \N(0,I),
$$
to finally return $x_T$.
In the backward pass, we start with randomly chosen $x_T \sim \N(0,I)$, and put
$
x_{t-1}=\mu(t,x_t)+\gamma_t \varepsilon_t,
$
where $\gamma_t$ are constants, $\mu$ is a deep network (typically given by U-NET), and $\varepsilon_t \sim \N(0,I)$.
The function $\mu$ and the constants $\gamma_t$ are chosen so that the trajectories of the forward and backward pass constructs cannot be distinguished.

After training, we can randomly sample data from the Gaussian distribution and use the denoising algorithm (backward pass) to achieve new samples. In practice, we can condition the U-NET model using a text prompt, which allows us to generate elements with arbitrary given properties. In addition, ControlNet \cite{zhang2023addingconditionalcontroltexttoimage} enables the denoising process to be conditioned by the image.  

\paragraph{\our{} }
Our model is a hybrid of the classical deepfake model and neural rendering. We can use NeRF or GS models. In practice, we can use any other approach to produce novel views from 2D images, such as NerFace~\cite{gafni2021dynamic}. In the end, avatars can be modified using a diffusion model guided by text prompts.

To unify the notation, we will treat the deep learning model for face manipulations as a black box that takes a set of images with camera positions:
$$
\I = \{(I_i, {\bf d}_i ) \}_{i=1}^{k},
$$
where $I_i$ is a 3D image from training data and ${\bf d}_i$ is the camera position of the $I_i$ image.  

We define a neural rendering model by the following function: 
$$
F({\bf d} ; \I , {\bf \Theta} ) = I_{ {\bf d} },
$$
that depends on training images with camera positions $\I$ and parameters ${\bf \Theta}$ that are adjusted during training. In the case of NeRF, this function describes the parameters of the neural network. On the other hand, the case of GS describes the parameters of Gaussian components. Our neural rendering process returns image $I_{ {\bf d} }$ from given position ${\bf d}$.

\begin{table*}[]
{\small
    
    \caption{
    We conduct a comparison using data from two 3D avatars. We select an image from the target avatar and employ it in the same manner as in the initial experiment. The remaining views of the target avatar are utilized as testing data. In this experiment, we evaluate the outcomes of \our{} compared to the original training views of the target avatar. As a reference point, we use a model that directly applies 2D deepfake on the training views of the target avatar.
    }
\begin{center}
    \begin{tabular}{l|llllllll}
    \multicolumn{8}{c}{PSNR $\uparrow$} \\
         & FACE 1 & FACE 2 & FACE 3 & FACE 4 & FACE 5 & FACE 6 & FACE 7 & Avg. \\ \hline
deepfake 2D & 12.75 & 11.97 & 13.94 & 11.35 & 11.47 & 12.31 & 11.84 & 12.23\\ 
NeRF & 9.73 & \bf 12.92 & 12.18 & \bf 11.76 & \bf 12.08 &\bf 12.91 & 9.73 & 11.62 \\ 
Gaussian Splatting & \bf 12.81 & 12.04 & \bf 13.86 & 11.42 & 11.51 & 12.33 & \bf 11.89 & \bf 12.27 \\  

 \hline

    \multicolumn{8}{c}{SSIM $\uparrow$} \\
\hline
deepfake 2D & 0.75 & 0.76 & 0.74 & 0.70 & 0.72 & 0.75 & 0.75 & 0.74\\ 
NeRF & \bf 0.86 & \bf 0.83 & \bf 0.81 & \bf 0.79 & \bf 0.82 & \bf 0.80 & \bf 0.86 & \bf 0.82\\ 
Gaussian Splatting & 0.75 & 0.76 & 0.75 & 0.71 & 0.72 & 0.75 & 0.75 & 0.74\\  
\hline

    \multicolumn{8}{c}{LPIPS $\downarrow$ }\\ \hline
deepfake 2D & 0.28 & 0.27 & 0.26 & 0.27 & 0.25 & 0.25 & \bf 0.24 & 0.26\\ 
NeRF & \bf 0.24 & \bf 0.26 & \bf 0.25 & \bf 0.26 & \bf 0.24 & \bf 0.24 & \bf 0.24 & \bf 0.25\\ 
Gaussian Splatting & 0.27 & 0.27 & 0.26 & 0.27 & 0.25 & 0.25 & \bf 0.24 & 0.26\\  
\hline
    \end{tabular}
    \end{center}
    \label{tab:nerf}
    }
\end{table*} 

We denote the deepfake model by: 
$$
\hat I_{s,t} = \G( I_{s} , I_{t}; \Phi ),  
$$
where $I_s$ and $I_t$ are the faces of the source and target images, respectively,  $\Phi$ denote all parameters of the deepfake network, in our case GHOST \cite{9851423}.  Our deepfake model returns the newly generated face $\hat I_{s,t}$.

In terms of diffusion modification, we denote the diffusion model by:
$$
\hat I_{s,t} = \G( I_{s} , I_{t}, \P; \Phi ),  
$$
where $I_s$ and $I_t$ are the faces of the source and target images, respectively, $\P$ is a text describing the modification (prompt), and $\Phi$ denotes all parameters of the diffusion network.

The general framework for our \our{} consists of two parts. First, we take training images of the 3D face model:
$$ 
\I = \{(I_i, {\bf d}_i ) \}_{i=1}^{k},
$$
and target image $I_{t}$ for the deepfake procedure (alternative modification based on diffusion).
Then, we apply the deepfake (alternative diffusion) model separately for all modified images:
$$
\hat \I = \{ \hat I_{i,t} \}_{i=1}^{k} = \{ \G( I_{i} , I_{t}; \Phi )\}_{i=1}^{k}.  
$$
Next, we train neural rendering on modified images to obtain a model dedicated to producing novel views $I_{ {\bf d} }$:
$$
F({\bf d} ; \hat \I , {\bf \Theta} ) = I_{ {\bf d} }.
$$

The above procedure is relatively simple and can generate plausible 3D deepfake avatars, as shown in Fig.~\ref{fig:deep_fake_GS}.




\begin{figure*}[th]
 	\centering
    \includegraphics[width=0.9\textwidth]{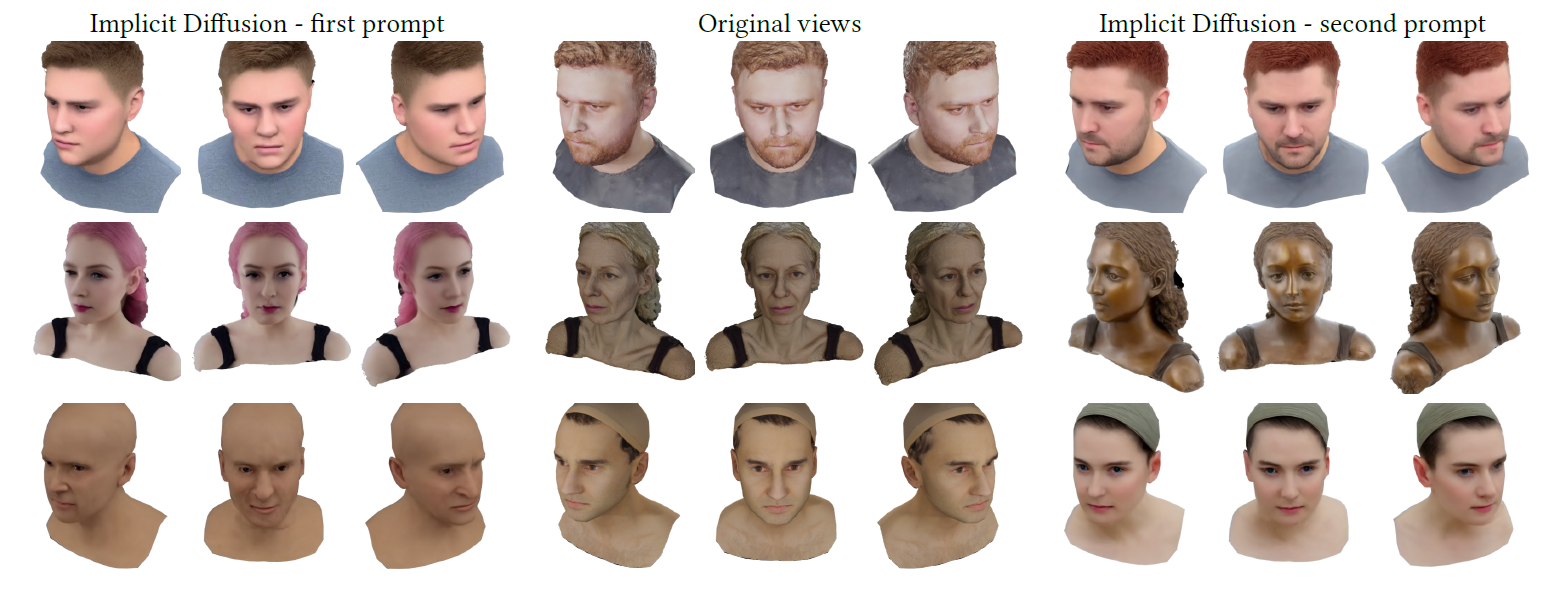} 

\caption{Results of Implicit Diffusion for two different faces. Each row shows the original avatar and two final 3D models generated using two different prompts.}
\label{fig:implicit_sd}
\end{figure*}

\begin{table}[]
{\small 
    \caption{
    Comparison between results obtained using deepfake 2D alone and \our{} with help of NeRFace. Results are calculated using faces from Fig.~\ref{fig:deepfake_nerface}. }
\begin{center}
    \begin{tabular}{l|lll}
    \multicolumn{4}{c}{PSNR $\uparrow$} \\
                    & FACE 1 & FACE 2 & FACE 3 \\ \hline
deepfake 2D         & \bf 25.12 & \bf 34.59 & \bf 27.25 \\         
NeRFace             & 20.33     & 28.43     & 25.50  \\

\hline

    \multicolumn{4}{c}{SSIM $\uparrow$} \\
\hline
deepfake 2D         & \bf 0.90 &  \bf 0.98 & \bf 0.91 \\
NeRFace             & 0.84     & 0.92      &  0.89 \\

\hline

    \multicolumn{4}{c}{LPIPS $\downarrow$ }\\ \hline
deepfake 2D         & \bf 0.11 & \bf 0.02 & \bf 0.10 \\
NeRFace             &  0.21    & 0.13     & 0.16\\
 
\hline
    \end{tabular}
    \end{center}
    \label{tab:nerface}
    }
\end{table}

\section{Experiments}

In what follows below, we present how our model works in practice. First, we present results on classical NeRF and GS responses, using \cite{liu2015faceattributes} as celebrity photos. Then, we show that our approach works with dynamic avatars. 

\paragraph{Static Deepfakes}

We start from a classical approach when we have one target image of a celebrity and the 3D object of a 3D human avatar. We aim to change the latter into a celebrity using a target image, as shown in Fig.~\ref{fig:deep_fake_GS}.

In our experiment, we apply classical 2D deepfake on source images. Then, we train our NeRF or GS model on modified images to obtain a 3D avatar of a given celebrity. To evaluate our model, we chose novel views of the source 3D avatar and applied deepfake. Such images were compared with the results generated by \our{}. The numerical comparison between NeRF and GS is presented in Tab.\ref{tab:celeb}. As we can see, GS obtained slightly better results.

In the second experiment, we wanted to contrast our results with the actual images of the target celebrity. Therefore, we used two 3D avatars instead of one. We take one image from the target avatar and use it the same way as in the first experiment. The other views of the target avatar are used as testing data. In such an experiment, we compare the results of \our{} with the original training views of the target avatar as shown in Fig.~\ref{fig:deep_fake_GS_1}. As a baseline, we used a model that directly applies 3D deepfake on training views of the target avatar. A numerical comparison between NeRF and GS can be seen in Tab.~\ref{tab:nerf}. Here the results of two proposed approaches are comparable in terms of quality.

\begin{figure*}[th]
	\centering
    \includegraphics[width=0.9\textwidth]{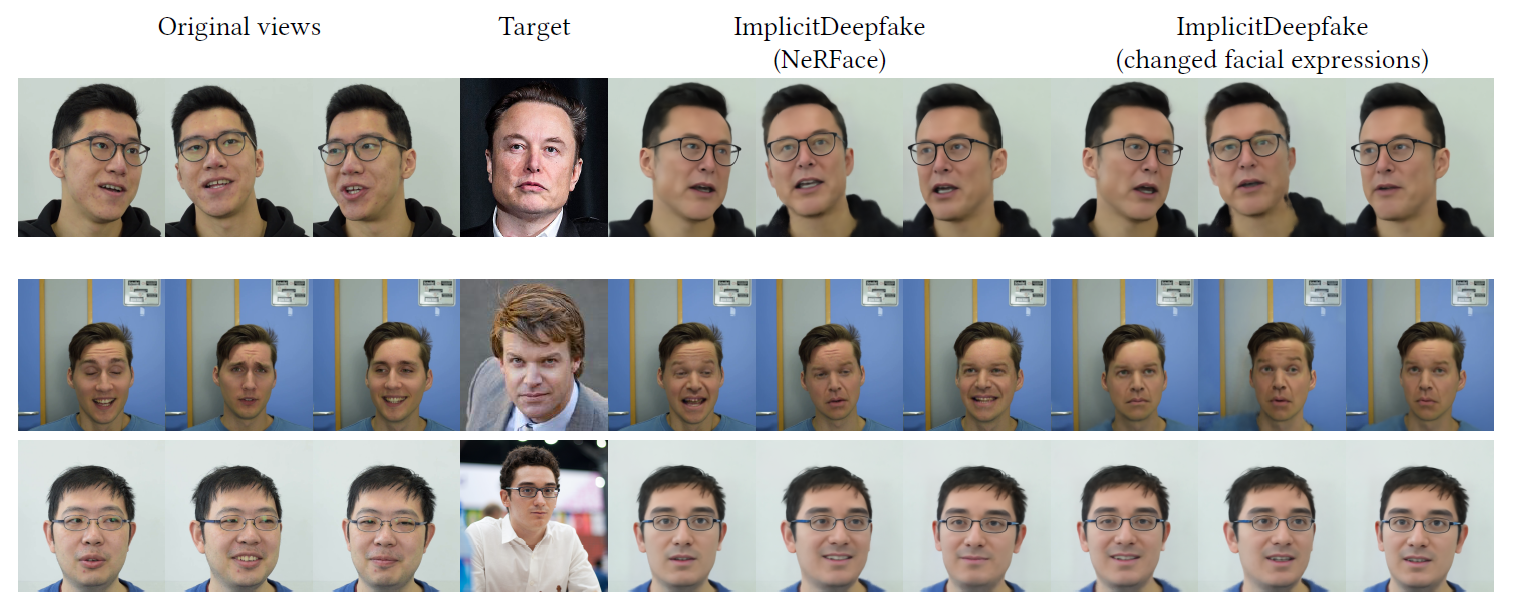} 

\caption{Results of \our{} training using NeRFace: original video frames, the target avatar image used for training, NeRFace-generated results, and the same images with altered facial expressions. }
\label{fig:deepfake_nerface} 
\end{figure*}

As a first conclusion, we can observe that applying 2D deepfake constructs well-aligned data. Furthermore, the independent modification produces consistent views for training both NeRF and GS. 

We can also see that GS gives sharper renders, as seen in Tab.~\ref{tab:celeb}. The difference is caused by NeRF, which may occasionally produce blurred renders. We think minor inconsistencies in the 2D deepfake model produce such effects.

In conclusion, we can infer that GS is more robust and consistent in dealing with issues related to the classical deepfake algorithm compared to NeRF.

\paragraph{Dynamic Avatars}

In practice, our model can be used in all generalizations of the NeRF and GS algorithms. To maximize such properties, we applied \our{} to the NerFace~\cite{gafni2021dynamic}, a 4D facial avatar of dynamic NeRFs that works directly on the video. 
To manage facial dynamics, NerFace integrates a scene representation network with a low-dimensional morphable model that allows precise manipulation of poses and expressions. It also employs volumetric rendering to produce images from this combined representation and shows that such a dynamic neural scene model can be learned solely from monocular input data, eliminating the requirement for a specialized capture system.  

In our experiments, we applied 2D deepfake techniques to the input video. We split the video into frames, processed through a 2D deepfake model. Subsequently, we trained the NeRFace model on the modified images to create a 4D avatar of a given celebrity, with the results shown in Fig.~\ref{fig:deepfake_nerface}. Our model also allows for manual control of facial expressions, as demonstrated in the right column.

We used a portion of the input video that was not included in the training data to evaluate our model. We calculated metrics such as PSNR, SSIM, and LPIPS between the original images and the images processed only through the 2D deepfake model and between the original images and the images processed through \our{}. The numerical comparison is presented in Tab.~\ref{tab:nerface}.


\begin{table*}[]
{\small 
    \caption{
    Comparison between results with two different 2D deepfake methods. Results are calculated using Gaussian Splatting on faces from Fig.~\ref{fig:deep_fake_GS}. }
\begin{center}
    \begin{tabular}{l|lllllllll}
    \multicolumn{9}{c}{PSNR $\uparrow$} \\
         & FACE 1 & FACE 2 & FACE 3 & FACE 4 & FACE 5 & FACE 6 & FACE 7 & FACE 8 & Avg. \\ \hline
GHOST & \bf 41.58 & \bf 41.71 & \bf 42.45 & \bf 44.00 & \bf 41.25 & \bf 41.73 & \bf 39.65 & \bf 41.43 & \bf 41.73\\ 
Rope & 38.52 & 39.18 & 39.54 & 42.71 & 37.78 & 39.34 & 38.14 & 38.29 & 39.19 \\ 
 \hline

    \multicolumn{9}{c}{SSIM $\uparrow$} \\
\hline
GHOST & \bf 0.99 &  \bf  0.99 &  \bf  0.99 &  \bf  0.99 &  \bf 0.99 & \bf 1.00 &  \bf 0.99 &  \bf 0.99 &  \bf 0.99\\
Rope & 0.97 & 0.97 & \bf 0.99 & 0.98 & 0.98 & 0.98 & 0.98 & 0.98 & 0.98\\ 

\hline

    \multicolumn{8}{c}{LPIPS $\downarrow$ }\\ \hline
GHOST &  \bf  0.04 &  \bf  0.04 &  \bf  0.03 &  \bf  0.03 &  \bf  0.02 &  \bf  0.01 &  \bf  0.02 &  \bf  0.02 & \bf  0.03 \\
Rope & 0.08 & 0.08 & 0.07 & 0.06 & 0.05 & 0.04 & 0.05 & 0.04 & 0.06 \\
 
\hline
    \end{tabular}
    \end{center}
    \label{tab:deepfakes}
    }
\end{table*}

\paragraph{Ablation Study of Deepfake}
To understand the impact of the choice of the 2D deepfake model on our pipeline, we conducted an ablation study by substituting the initial 2D deepfake model. We experimented with models such as GHOST and rope\footnote{url{https://github.com/Hillobar/Rope}}. Metrics such as PSNR, SSIM, and LPIPS were calculated to quantitatively compare the performance of each configuration. Our results, presented in Tab.~\ref{tab:deepfakes}, indicate that, although there were slight differences, both models could generate realistic avatars with comparable visual quality and structural fidelity.

\paragraph{Diffusion Model}

When we have a 3D avatar of the face, we can apply additional modification by using simple text prompts. We can do this by diffusion models. In our model, we use a method that leverages the power of Stable Diffusion \cite{DBLP:journals/corr/abs-2112-10752}, ControlNet \cite{zhang2023addingconditionalcontroltexttoimage} and Example-Based Synthesis (EbSynth) \cite{Jamriska19-SIG} to apply transformations in a coherent and visually appealing manner.

The first phase of such a pipeline involves generating sequential render images of the original avatar. From this sequence, a subset of frames at regular intervals is selected. These frames are then processed using the Stable Diffusion model \cite{DBLP:journals/corr/abs-2112-10752}. All selected frames are simultaneously placed on the same input image when applying the prompt to ensure consistency across transformations. During this process, ControlNet \cite{zhang2023addingconditionalcontroltexttoimage} adjusts the Stable Diffusion model, preserving structural integrity by conditioning the generated images on the edges of the original frame.

In the second phase, EbSynth \cite{Jamriska19-SIG} propagates the transformations to the remaining sequence frames. EbSynth uses optical flow calculations to ensure temporal coherence and stylistic consistency across all frames.

In the final phase, we used GS to reconstruct a 3D model from transformed frames. This process integrates the modified visual information into a 3D representation, allowing for novel view synthesis and high-quality avatar generation.

By structuring the pipeline in this manner, advanced visual processing techniques are seamlessly integrated, ensuring the high quality and versatility of the resulting avatars.
The results of diffusion modification are presented in Fig.~\ref{fig:implicit_sd}

\section{Conclusions}

This paper introduces \emph{\our{}}, a method that produces a human face avatar from a single image that utilizes the conventional 2D deepfake or 2D diffusion model to alter the training images for NeRF or GS. Subsequently, NeRF and GS are trained separately on the modified facial images, generating realistic and plausible deepfakes that can serve as 3D avatars. Compared to NeRF, GS produces sharper renders, with the former occasionally generating blurred renders. We attribute these effects to minor inconsistencies introduced by the 2D deepfake model. In general, GS demonstrates greater resilience to view inconsistencies than NeRF.

\paragraph{Societal impact}

Deepfake technology offering the ability to take actions using a public figure's personality is a serious threat that will increase along with the advent of new breakthroughs in the field of computer vision and shall never be underestimated. At the same time, modeling of character faces comes with multiple emerging and underexplored practical applications that create a base for educational, entertaining, and development purposes. The latter include the creation of persons in human-computer interaction research or the provision of an interface for artificial intelligence agents. Furthermore, \our{} can be used in the entertainment industry as an alternative to existing video game avatar creation.


\end{document}